\documentclass[12pt,journal,onecolumn,draftclsnofoot]{IEEEtran}
\usepackage[utf8]{inputenc} % allow utf-8 input
\usepackage[T1]{fontenc}    % use 8-bit T1 fonts
\usepackage{xcolor}         % colors
\usepackage{thm-restate}
\usepackage{soul}
\usepackage{ifthen}
\usepackage{cite}
\usepackage{amsmath} 
\usepackage{url,color,cite}
\usepackage{caption}
\usepackage{amsmath,amsthm,amsfonts,amssymb}
\usepackage{cases}
\usepackage{mathtools}
\DeclarePairedDelimiter{\ceil}{\lceil}{\rceil}
\DeclarePairedDelimiter{\floor}{\lfloor}{\rfloor}
\usepackage{float}
\usepackage{epstopdf}
\usepackage{nicefrac}
\usepackage{algorithm}
\usepackage[noend]{algpseudocode}
\usepackage{pifont}
\usepackage{subcaption}
\usepackage{sidecap}
\usepackage{multirow}
\usepackage{multicol}
\usepackage{bbm}

\newtheorem{assumption}{Assumption}

\newtheorem*{lemma*}{Lemma}
\newtheorem{theorem}{Theorem}
\newtheorem*{theorem*}{Theorem}
\theoremstyle{definition}

\theoremstyle{definition}

\newtheorem{remark}{Remark}
\theoremstyle{definition}
\newtheorem{corollary}{Corollary}
\theoremstyle{definition}

\newtheorem*{claim*}{Claim}

\newcommand{\thetas}{\theta_\star}
\newcommand{\thetat}{\hat{\theta}_t}
\newcommand{\rt}{\hat{r}_t}

\newcommand{\Xt}{\hat{X}_t}
\newcommand{\ri}{\hat{r}_i}
\newcommand{\Xhi}{\hat{X}_i}
\newcommand{\etai}{\hat{\eta}_i}
\newcommand{\thetai}{\hat{\theta}_i}

\usepackage{microtype}
\usepackage{graphicx}
\usepackage{booktabs} 
\usepackage{hyperref}
\usepackage{cleveref}
\usepackage{enumitem}

\usepackage{lipsum}

\begin{document}

\title{Learning in Distributed Contextual Linear Bandits Without Sharing the Context} 

\author{\IEEEauthorblockN{Osama A. Hanna$^\dagger$, Lin F. Yang$^\dagger$ and Christina Fragouli$^\dagger$\\ 
$^\dagger$University of California, Los Angeles\\
Email:\{ohanna, linyang, christina.fragouli\}@ucla.edu}
}
\maketitle

\begin{abstract}
	Contextual linear bandits is a rich and theoretically important model that has many practical applications. Recently, this setup gained a lot of interest in applications over wireless where communication constraints can be a performance bottleneck, especially when the contexts come from a large $d$-dimensional space. In this paper, we consider a distributed memoryless contextual linear bandit learning problem, where the agents who observe the contexts and take actions are geographically separated from the learner who performs the learning while not seeing the contexts. We assume that contexts are generated from a distribution and propose a method that uses $\approx 5d$ bits per context for the case of unknown context distribution and $0$ bits per context if the context distribution is known, while achieving nearly the same regret bound as if the contexts were directly observable. 
The former bound improves upon existing bounds by a $\log(T)$ factor, where $T$ is the length of the horizon, while the latter achieves information theoretical tightness.
\end{abstract}

{\allowdisplaybreaks

\section{Introduction} \label{sec:intro}

Contextual linear bandits offer a sequential decision-making framework that
combines fundamental theoretical importance with  significant practical popularity \cite{auer2002using}, as it offers a tractable way to capture side information (context), as well as a potentially infinite set of decisions (actions). The most prominent application is in recommendation systems \cite{mary2015bandits}, but it has also been used in applications such as virtual support agents \cite{sajeev2021contextual}, %\cite{Sajeev2021},
clinical trials \cite{durand2018contextual}, transportation systems \cite{awerbuch2004adaptive},
%\cite{transportation1,transportation2},
wireless optimization \cite{li2020multi, le2014sequential},
%\cite{wireless1},
health \cite{bouneffouf2017bandit}, robotics \cite{matikainen2013multi} and online education \cite{rafferty2018bandit}.

In this paper, we develop algorithms that support the deployment of contextual linear bandits in distributed settings. In particular, we consider the case where a central learner wishes to solve a contextual linear bandit problem with the help of transient agents. That is, we assume that the agents do not keep memory of past actions and may not be present for the whole duration of learning; learning in our setup  can happen thanks to the persistent presence of the central learner.
We view the central learner as a ``knowledge repository'', that accumulates knowledge from the experience of the transient agents and makes it available to next agents. The central agent, through the information it keeps, could help  passing by devices decide how to perform an action, for example:  passing by drones decide how to perform a manoeuver; agricultural robots decide what amounts of substances such as pesticids to release; and passing by mobile devices decide which local restaurants to recommend.

The main challenge we try to address in this paper is the efficient communication of the context the agents experience.
More specifically,  in our setup, each time an agent joins, she receives from the central learner information on the system, such as current estimates of the system parameters; she observes her current context, selects and plays an action  and collects the corresponding reward.  Note that although the distributed agent knows her context, the action she plays and the observed reward, the central learner does not - and needs this information to update its estimate of the system parameters. The context in particular can be communication heavy - in the examples we mentioned before, for drones the context could be their navigation capabilities, physical attributes, and enviromental factors such as wind speed; for agricultural robots, it could be images that indicate state of plants and sensor measurements such as of soil consistency; for restaurant recommendations, it could be the personal dietary preferences and restrictions, budget, and emotional state. Moreover, because of geographical separation, the central agent may not have any other way to learn the context beyond communication. Unlike the reward, that is usually a single scalar value, the context can be a vector of a large dimension $d$ from an infinite alphabet, and thus, communicating the context efficiently is heavily nontrivial.

The technical question we ask is, \emph{how many bits do we need to convey per context to solve the  linear bandit problem without downgrading the performance as compared to the non-distributed setting?}

%The technical question we ask is, \emph{how many bits do we need to convey to the central learner per context to solve the contextual linear bandit problem, without downgrading the learning performance as compared to the non-distributed setting?}

In this paper, we design algorithms that support this goal. We note that our algorithms  optimize the uplink communication (from the agents to the learner), and assume unlimited (cost-free) downlink communication. This is a standard assumption in wireless \cite{anisi2015survey, novlan2013analytical, hanna2022solving} for several reasons: uplink wireless links tend to be much more bandwidth restricted, since several users may be sharing the same channel; uplink communication may also be battery-powered and thus more expensive to sustain; in our particular case, the agents may have less incentive to communicate (provide their feedback) than the learner (who needs to learn). Having said that, 
we note that our  algorithms (in Sections~\ref{sec:0bits} and~\ref{sec:dbits}) make frugal use of the downlink channels, only using them to transmit  system parameters.
%But again this is not something we optimize for, in this paper we focus on minimizing the number of bits the agents need to send to the central learner.

Below we summarize our main contributions:\\
% (with the same order regret) as if the central agent had access to all the information the agents do (full precision unquantized communication). We assume in this work that contexts are generated from a distribution, that may be known or unknown.  Our main contributions are as follows. 
%\textcolor{cyan}{To update, eg if we say something about approximations}
%\begin{enumerate}
%\item
{\bf   1.} We show the surprising result that, if the central agent knows the distribution of the contexts,  we do not need to communicate the context at all - the agent does not need to send any information on the actual context she observes and the action she plays. 
It is sufficient for the agent to  just send $1$ bit to convey quantized information on her observed reward and nothing else.
But for this very limited communication, the central learner can learn a policy that achieves the same order of regret  as if full information about the context and reward is received.
%as when receiving in full precision all the information that the agents have.
This result holds for nearly all context distributions %(under some mild common conditions), 
and it is the best we can hope for - \emph{zero bits} of communication for the context.\\
%  {Note that only adding a small amount of zero mean noise to the true context by the agent before sending it to the learner can severely affect the regret in some cases~\cite{}.}
%Interestingly, this performance can be achieved using a polynomial time algorithm, that we term ****
{\bf 2.} If the central agent has no knowledge of the context distribution, we show that  $\approx 5d$ bits per context (where $d$ is the context dimension) is sufficient to achieve the same order regret as knowing the context in full precision. 
Note that previous algorithms use $O(d\log T)$ bits per context to achieve the same order regret, where $T$ is the length of the horizon \cite{lattimore2020bandit}, {and  require exponential complexity}. %\textcolor{cyan}{Please give a reference or refer to some section.}
%This can also be achieved using a polynomial time algorithm, that we term ***
%\end{enumerate}

{\bf Related Work and Distinction.} Contextual linear bandits is a rich and  important model that has attracted significant interest both in theory and applications \cite{auer2002using, lattimore2020bandit}. {Popular algorithms for this setup include LinUCB \cite{abbasi2011improved, rusmevichientong2010linearly} and cotextual Thompson sampling \cite{agrawal2013thompson}. Under Assumption~\ref{assump:1}, these algorithms achieve a regret of $O(d\sqrt{T\log T})$, where $d$ is the dimension of an unknown system parameter and $T$ is the time horizon, while the best known lower bound for this setup is $\Omega(d\sqrt{T})$ \cite{rusmevichientong2010linearly}. These algorithms assume perfect knowledge of the contexts and rewards.} Our work focuses on operation under communications constraints in a distributed setting, where the contexts and rewards are observed by a remote agent that communicates with the learner over a communication constrained channel.

There is large body of work focusing on distributed linear contextual bandits settings,
but mainly within the framework of federated learning, where batched algorithms have been proposed for communication efficiency \cite{shi2021federated,shahrampour2017multi, anantharam1987asymptotically, anandkumar2011distributed, landgren2019distributed} that aggregate together observations and parameter learning across a large number of iterations. This is possible because
in  federated learning, the agents themselves wish to learn the system parameters, remain active playing multiple actions throughout the learning process, and exchange information with the goal of speeding up their learning \cite{shi2021federated, shahrampour2017multi}.  
In contrast, in our setup batched algorithms cannot reduce the communication cost because each agent only plays a single action; this may be because agents are transient, but also because they may not be interested in learning - this may not be a task that the agents wish to consistently perform - and thus do not wish  to devote resources to it. For example, an agent  may wish to try a restaurant in a special occasion,
but would not be interested in sampling multiple restaurants/learning recommendation system parameters. %Thus
% These algorithms cannot be applied in our setup, as we assume that agents are transient and do not remain engaged with the system for more than one action. 
Our setup supports a different (and complementary) set of applications than the federated learning framework, and requires a new set of algorithms that operate without requiring agents to keep memory of past actions.\footnote{Our techniques could be adapted to  additionally improve the communication efficiency of batched algorithms, but this is not the focus of our work.}

There is a long line of research on compression for machine learning and distributed optimization, e.g., compression for distributed gradient descent \cite{seide20141, alistarh2017qsgd, mayekar2020ratq, hanna2021quantization}, and distributed inference \cite{hanna2020distributed}. However, such schemes are not optimized for active learning applications. Our compression schemes can be seen as quantization schemes for contexts and rewards tailored to active learning applications.

Our work also differs from traditional vector compression schemes \cite{gersho2012vector} that aim to reconstruct the data potentially with some distortion (achieve rate-distortion trade-offs). In our case, we do not aim to reconstruct the data, but instead to distinguish the best arm for each context. Indeed, using $0$ bits, as we do in Section~\ref{sec:0bits}, we cannot reconstruct a meaningful estimate of the context.

To the best of our knowledge, our framework has not been examined before for linear contextual bandits.
Work in the literature has examined compression for distributed memoryless MABs  \cite{hanna2022solving}, but only for  rewards (scalar values) and not the contexts (large vectors), and thus these techniques also  do not extend to our case.

{\bf Paper organization.}
%The rest of the paper is organized as follows.
Section~\ref{sec:notation} reviews our notation and problem formulation; Section~\ref{sec:0bits} provides and analyzes our algorithm for known  and Section~\ref{sec:dbits}  for unknown context distributions.

\section{Notation and Problem Formulation} \label{sec:notation}
%==============================================================

\textbf{Notation.} We use the following notation throughout the paper. For a vector $X$ we use $X_i$ or $(X)_i$ to denote the $i$-th element of the vector $X$; similarly for a matrix $V$, we use $V_{ij}$ or $(V)_{ij}$ to denote the element at row $i$, and column $j$. {We  use $\|V\|_2$ to denote the matrix spectral norm.} For a function $f$, we denote its domain and range by dom$(f)$, ran$(f)$ respectively. When dom$(f)\subseteq \mathbb{R}$, we use $f(X)$ for a vector $X\in \mathbb{R}^d$ to denote $f(X):=[f(X_1),...,f(X_d)]$, i.e., the function $f$ is applied element-wise; for example we use $X^2$ to denote the element-wise square of $X$. We denote the inverse of a function $f$ by $f^{-1}$; if $f$ is not one-to-one, with abuse of notation we use $f^{-1}$ to denote a function that satisfies $f(f^{-1}(x))= x \forall x\in \text{ran}(f)$ (this is justified due to the axiom of choice \cite{jech2008axiom}). For a matrix $V$, we use $V^{-1}$ to denote its inverse; if $V$ is singular, we use $V^{-1}$ to denote its pseudo-inverse. We use $[N]$ for $N\in \mathbb{N}$ to denote $\{1,...,N\}$, and $\{X_a\}_{a\in \mathcal{A}}$ to denote the set $\{(a,X_a)|a\in \mathcal{A}$\}. We say that $y=O(f(x))$ if there is $x_0$ and a constant $C$ such that $y\leq Cf(x)$ $\forall x>x_0$; we also use $\tilde{O}(f(x))$ to omit $\log$ factors.

\textbf{Contextual Linear Bandits.}
We consider a contextual linear bandits problem over a horizon of length $T$ \cite{auer2002using}, where at each iteration $t=1,...,T$, an agent, taking into account the context, chooses an action $a_t\in \mathcal{A}$ and receives a reward $r_t$.  For each action  $a\in \mathcal{A}$, the agent has access to a corresponding feature vector $X_{t,a}\in \mathbb{R}^d$. The  set of all such vectors  $\{X_{t,a}\}_{a\in \mathcal{A}}$  is the context  at time $t$, and the agent can use it to decide which action $a_t$ to play. We assume that the context is generated from a distribution, i.e., given $a$, $X_{t,a}$ is generated from a distribution $\mathcal{P}_a$. 
As a specific example, we could have that $a\in\mathbb{R}^d$ and $X_{t,a}$ is generated from a Gaussian distribution with zero mean and covariance matrix $||a||_2I$, where $I$ is the identity matrix, i.e., $\mathcal{P}_a=\mathcal{N}(0,||a||_2I).$ 
The selection of $a_t$ may depend not only on the current context $\{X_{t,a}\}_{a\in \mathcal{A}}$ but also on the history $H_{t}\triangleq \{ \{X_{1,a}\}_{a\in \mathcal{A}},a_1,r_1,...,\{X_{t-1,a}\}_{a\in \mathcal{A}},a_{t-1},r_{t-1} \}$, namely, all
previously selected actions, observed contexts and rewards. 
Once an action is selected, the reward is generated according to
\begin{equation}\label{eq:rew}
	r_t=\langle X_{t,a_t},\theta_\star \rangle+\eta_t,
\end{equation} 
where $\langle .,.  \rangle$ denotes the dot product, $\theta_\star$ is an unknown (but fixed) parameter vector in $\mathbb{R}^d$, and $\eta_t$ is noise.
We assume that the noise follows an unknown distribution with $\mathbb{E}[\eta_t|\mathcal{F}_t]=0$ and $\mathbb{E}[\exp(\lambda \eta_t)|\mathcal{F}_t]\leq \exp(\lambda^2/2)\forall \lambda\in \mathbb{R}$, where  $\mathcal{F}_t=\sigma(\{X_{1,a}\}_{a\in \mathcal{A}},a_1,r_1,...,\{X_{t,a}\}_{a\in \mathcal{A}},a_t)$ is the filtration \cite{durrett2019probability} of historic information up to time $t$, and 
%\textcolor{blue}{(it depends also of the context and action at time $t$? so $\eta_t$ depends on $a_t$? or can we just write $H_t$?)} \textcolor{red}{OH: linear bandit algorithms work even when $\eta_t$ is a function of $a_t$. We also need this because the quantization error can be dependent on $a_t$.},
$\sigma(X)$ is the $\sigma$-algebra generated by $X$ \cite{durrett2019probability}.

The  objective is to minimize the regret $R_T$ over a horizon of length $T$, where
\begin{equation}
	R_T = \sum_{t=1}^T \max_{a\in \mathcal{A}} \langle X_{t,a}, \theta_\star \rangle - \langle X_{t,a_t}, \theta_\star \rangle.
\end{equation}
The best performing algorithms for this problem, such as LinUCB \cite{abbasi2011improved, rusmevichientong2010linearly} and contextual Thompson sampling \cite{agrawal2013thompson}, achieve a worst case regret of $O(d\sqrt{T\log T})$ \cite{lattimore2020bandit}. The best known lower bound is $\Omega (d\sqrt{T})$ \cite{rusmevichientong2010linearly}.
%\textcolor{blue}{Please add references.}

In the rest of this paper, we make the following  assumptions that are standard in the literature \cite{lattimore2020bandit}.
\begin{assumption}\label{assump:1}
	We consider contextual linear bandits that satisfy:\\
	%	\begin{enumerate}
		%\item
		{\bf (1.)} $\|X_{t,a}\|_2\leq 1$, $\forall t\in [T]$, $a\in \mathcal{A}$.  \qquad
		%		\item
		{\bf (2.)}  $\|\thetas\|_2\leq 1$. \qquad
		%		\item
		{\bf (3.)} $r_t\in [0,1]$, $\forall t\in [T]$.
		%	\end{enumerate}
\end{assumption}

\noindent\textbf{Memoryless Distributed Contextual Linear Bandits.} We consider a distributed setting that consists of a central learner communicating with geographically separated agents. For example, the agents are drones that interact with a traffic policeman (central learner) as they fly by. We assume that the agents do not keep memory of past actions and may not be present for the whole duration of learning; learning in our setup  can happen thanks to the persistent presence of the central learner.

At each time $t$, $t=1\ldots T$, a distributed agent joins the system; she receives from the central learner information on the system, such as the current estimate of the parameter vector $\theta_\star$ or the history $H_t$;  she observes the current context $\{X_{t,a}\}_{a\in \mathcal{A}}$, selects and plays an action $a_t$ and collects the corresponding reward $r_t$.  
Note that although the distributed agent knows the context $\{X_{t,a}\}_{a\in \mathcal{A}}$, the action $a_t$ and the observed reward $r_t$, the central learner does not. The central learner may need this information to update its estimate of the
system parameters, such as the unknown parameter vector $\theta_*$, and the history $H_{t+1}$. 
However, we assume that the agent is restricted to utilize a communication-constrained channel and thus may not be able to send the full information to the central learner. %needs, unquantized.%, namely, %can only  one message $M_t$ using $B_t$ bits.

The main question we ask in this paper is: can we design a compression scheme,
where the agent sends to the learner only one message using $B_t$ bits (for as small as possible a value of $B_t$) that enables the central learner to learn equally well (experience the same order of regret) as if there were no communication constraints? With no communication constraints the agent could send unquantized the full information $\{ \{X_{t,a}\}_{a\in \mathcal{A}}, a_t, r_t\}$.
Instead, the agent transmits a message that could be  a function of all locally available information at the agent. For example, it could be a function of $(H_t,\{X_{t,a_t}\}_{a\in \mathcal{A}},a_t,r_t)$, if the agent had received $H_t$ from the learner. It could also be a function of just $(X_{t,a_t},r_t)$, which could be sufficient if the central learner employs an algorithm such as LinUCB \cite{abbasi2011improved, rusmevichientong2010linearly}.
In summary, we set the following goal. 

\textbf{Goal.} Design contextual linear bandit schemes for the memoryless distributed setting that achieve the best known regret of $O(d\sqrt{T\log(T)})$, while communicating a small number of bits $B_t$.

%As motivated in Section~\ref{sec:intro},
We only impose communication constraints on the uplink communication (from the agents to the central learner) and assume no cost downlink communication (see discussion in Secttion~\ref{sec:intro}).
%The technical question we ask is, \emph{how many bits do we need to convey to the central learner per context to solve the contextual linear bandit problem, without downgrading the learning performance as compared to the non-distributed setting?
	%; The central learner can send information to the agents (downlink) at no cost.%\lin{remember to add discussion on this in the intro.} 
	
	\iffalse
	. The central node then updates $H_t$ to obtain $H_{t+1}$. The purpose of sharing information about the history $H_t$ is to help the distributed user to pick a good action and potentially improve the users compression scheme. We assume that users immediately communicate the message $M_t$, i.e., no memory at the users side. This allows to support applications where users are in the system for only one (or a few) time slot, and applications with simple users that do not have enough memory, or are not willing to allocate memory, to keep information about the history. While we care about minimizing the rate of the messages $M_t$ (uplink communication), we do not count the transmissions from the central learner to the users (downlink) in the communication cost. This is because unlike the users, the learner can have no power restrictions and also due to the fact that uplink is usually a much more congested channel. 
	\fi

	\textbf{Stochastic Quantizer (SQ) \cite{gray1993dithered}.} Our proposed algorithms use stochastic quantization, that we next review. We define SQ$_{\ell}, \ell \in \mathbb{N}$ to be a quantizer, that uses $\log(\ell+1)$ bits, consisting of an encoder and decoder described as following. The encoder $\xi_{\ell}$ takes a value $x\in [0,\ell]$ and outputs an integer value
	\begin{equation} \label{eq3}
		\xi_{\ell}=\left\{
		\begin{array}{ll}
			\floor{x}  & \mbox{with probability } \ceil{x}-x \\
			\ceil{x} & \mbox{with probability } x-\floor{x}.
		\end{array}
		\right.
	\end{equation}
	The output $\xi_{\ell}$ is represented with $\log(\ell+1)$ bits. The decoder $D_{\ell}$ takes as input the binary representation of $\xi_{\ell}(x)$ and outputs the real value $\xi_{\ell}(x)$. The composition of the encoder $\xi_\ell$, the binary mapping, and decoder $D_\ell$ is denoted by SQ$_\ell$. We notice that since the decoder only inverts the binary mapping operation, we have that SQ$_\ell=\xi_\ell$. When SQ$_\ell$ is applied at the agents side, the agent encodes its data, $x$, as $\xi_\ell(x)$, then sends the corresponding binary mapping to the central learner that applies $D_\ell$ to get SQ$_\ell(x)$. With slightly abuse of notation, this operation is described in the paper, by saying that the agent sends SQ$_\ell$ to the central learner.
	
	The quantizer SQ$_\ell$ is a form of dithering \cite{gray1993dithered} and it has the following properties
	\begin{align*}
		&\mathbb{E}[\text{SQ}_\ell(x)|x] = \floor{x}(\ceil{x}-x)+\ceil{x}(x-\floor{x})=x(\ceil{x}-\floor{x})=x, \quad\text{and}\quad
		&|\text{SQ}_\ell(x)-x|\leq 1.
	\end{align*}
	In particular, it conveys an unbiased estimate of the input with a difference that is bounded by $1$ almost surely. We also define a generalization of SQ$_\ell$ denoted by SQ$_\ell^{[a,b]}$ where the input $x$ of the encoder is in $[a,b]$ instead of $[0,\ell]$. The encoder first shifts and scales $x$ using $\tilde{x}=\frac{\ell}{b-a}(x-a)$ to make it lie in $[0,\ell]$, then feeds $\tilde{x}$ to the encoder in \eqref{eq3}. This operation is inverted at the decoder. It is easy to see that SQ$_\ell^{[a,b]}$ satisfies
	\begin{align*}
		\mathbb{E}[\text{SQ}_\ell^{[a,b]}(x)|x] =x, |\text{SQ}_\ell^{[a,b]}(x)-x|\leq \frac{b-a}{\ell}.
	\end{align*}

\section{Contextual Linear Bandits with Known Context Distribution} \label{sec:0bits}
%==================================================================
In this section, we show that if the central learner knows the distributions for the vectors $X_{t,a}$, then the agent does not need to convey the specific realization of the vector   $X_{t,a}$ she observes  at all - it is sufficient to just send $1$ bit to convey some information on the observed reward and nothing else.  But for this very limited communication, the central learner can experience the same order regret, as when receiving in full precision all the information that the agents have, namely,   $R_T=O(d\sqrt{T\log T})$.  Algorithm~\ref{alg:known}, that we describe in this section, provides a method to achieve this.  Algorithm~\ref{alg:known} is clearly optimal, as we cannot hope to use less than zero bits for the vector $X_{t,a}$.

\begin{remark}
	Knowledge of the distribution of  $X_{t,a}$ is possible in practice, since many times the context may be capturing well studied statistics (e.g., male or female, age, weight, income, race, dietary restrictions, emotional state, etc) - the advent of large data has made and will continue to make such distributions available. Similarly, actions may be finite (eg.,  restaurants to visit) or well described (e.g., released amounts of substances), and thus the distribution of $X_{t,a}$ could be derived. When the distribution is approximately known, we provide later in this section a bound on the  misspefication performance penalty in terms of regret.
\end{remark}
%\textcolor{blue}{I do not know what are practical examples of the function that $X_{t,a}$ implements...}
%\lin{also need to discuss misspecification settings.}
%\textcolor{cyan}{If we have time to add misspecification we will refer to it here.} 
% Note that the distribution ....

%\textcolor{cyan}{CF: Lin, we rewrote big parts of the main idea/rest of this section, could you please look at it again?}\\
{\bf Main Idea.}
The intuition behind  Algorithm~\ref{alg:known}  is that it reduces the multi-context linear bandit problem to a single context problem.
In particular, it calls as a subroutine 
an algorithm we term $\Lambda$, that serves as a placeholder for any current (or future) bandit algorithm that achieves regret $O(d\sqrt{T\log T})$ for the case of a single context (for example, LinUCB \cite{abbasi2011improved, rusmevichientong2010linearly}). The central learner uses $\Lambda$ to convey to the agents the information they need to select a good action. Our aim is to parametrize  the single context problem appropriately, so that, by solving it we also solve our original problem.

Recall that in a single context problem, at each iteration $t$, any standard linear bandit algorithm $\Lambda$ selects a feature vector (an action) $x_t$ from a set of allowable actions $\mathcal{X}$, and observes a reward 
\begin{equation}\label{eq:rew_single}
	r_t=\langle x_t,\thetas(\Lambda) \rangle+\eta_t,	
\end{equation}	
where $\thetas(\Lambda)$ is an uknown parameter vector  and $\eta_t$ is noise that satisfies the same assumptions as in \eqref{eq:rew}.  The objective of $\Lambda$ is to minimize the standard linear regret expression $R_T(\Lambda)$ over a horizon of length $T$, namely
\begin{equation} \label{eq9}
	R_T(\Lambda)=\sum_{t=1}^T \max_{x\in \mathcal{X}}\langle x,\thetas(\Lambda)  \rangle - \langle x_t,\thetas(\Lambda) \rangle.
\end{equation}
Our reduction proceeds as follows. We assume that $\Lambda$ operates over the same horizon of length $T$ and is parametrized by an unknown parameter $\thetas(\Lambda)$.  
We  will design the action set  $\mathcal{X}$ that we provide to $\Lambda$  using our knowledge of the distributions $\mathcal{P}_a$\footnote{Recall that  given $a$, $X_{t,a}$ is generated from  distribution $\mathcal{P}_a$, see Section~\ref{sec:notation}.}  as we will describe later in (\ref{eq12}). During each iteration, the central learner asks $\Lambda$ to select an action $x_t \in \mathcal{X}$ and then provides to $\Lambda$ a reward for this action (our design  ensures that this reward satisfies \eqref{eq:rew_single} {with $\theta_\star(\Lambda)=\thetas$}). $\Lambda$ operates with this information, oblivious to what else the central learner  does. Yet, the action $x_t$ is never actually played:
the central learner uses the selected action $x_t$ to create an updated estimate of the parameter vector 
$\hat{\theta}_t$, as we will describe later, and only sends this parameter vector estimate to the distributed agent. The agent observes her context, selects what action to play, and sends back her observed quantized reward to the central learner. This is the reward that the central learner provides to $\Lambda$.
We   design the set $\mathcal{X}$ and the agent operation to satisfy that: \eqref{eq:rew_single} holds; and $R_T-R_T(\Lambda)$ is small, where $R_T$ is the regret for our original multi-context problem and $R_T(\Lambda)$ the regret of $\Lambda$.
We next try to provide some intuition on how we achieve this.

We first describe how we construct the set $\mathcal{X}$. Let $\Theta$ be the set of all values that $\thetas$ could possibly take.
For each possible parameter vector value $\theta \in \Theta$ the central learner  considers the quantity
\begin{equation} \label{eq10}
	X^{\star}(\theta) = \mathbb{E}_{\{x_a: x_a\sim P_a\}}[\arg\max_{x\in\{x_a: a\in A\}} \langle x, \theta \rangle ]
\end{equation}
where $x_a$ is the random variable that follows the distribution $\mathcal{P}_a$.
Note that the function $X^{\star}:\mathbb{R}^d\to \mathbb{R}^d$  can be computed offline before the learning starts, see Example~1. We then use
\begin{equation}\label{eq12}
	\mathcal{X}=\{X^\star(\theta)| \theta\in \Theta\}.
\end{equation}
Intuitively, for each value of $\theta$, we optimistically assume that the distributed agent may select the best possible realization $X_{t,a}$ for this $\theta$ (that has the expectation in \eqref{eq10}),
and receive the associated reward; accordingly, we restrict the action space $\mathcal{X}$ of $\Lambda$ to only contain the expectation of these ``best''  $X_{t,a}$.
The vector $x_t\in \mathcal{X}$ may not actually be the vector corresponding to the action the agent selects; it is only used to convey to the agent an estimate of the unknown parameter $\thetat$ that satisfies $x_t=X^\star(\thetat)$. Although the learner does not control which action the agent plays, this is influenced by $\thetat$; we show in App.~\ref{app:known} that $X_{t,a_t}$ is an unbiased estimate of $x_t$, and the generated reward follows the linear model in \eqref{eq:rew_single} with $\thetas(\Lambda)=\thetas$. In Theorem~\ref{thm:known}, we prove that 
\begin{equation}
	\arg\max_{x\in \mathcal{X}}\langle x,\thetas \rangle=X^\star(\thetas).
\end{equation}
Hence, if $\Lambda$ converges to selecting the best action for the single context problem, we will have that $\thetat=\thetas$. If  there are multiple values for $\theta$ with $X^\star(\theta)=X^\star(\thetas)$, we show in the proof of Theorem~\ref{thm:known} that they all lead to the same expected reward for the original multi-context problem. 

{\bf Example 1. }{Consider the case where $d=1$, $\mathcal{A}=\{1,2\}$, $X_{t,a}\in \{-1,1\}$ $\forall a\in \mathcal{A}$, $\Theta=\{-1,1\}$, $\thetas=1$ and $X_{t,1}$ takes the value $-1$ with probability $p$ and $1$ otherwise, while $X_{t,2}$ takes the values $-1$ with probability $q$ and $1$ otherwise. Then, we have that
	\begin{equation}
		\arg\max_{X_{t,a}} \langle X_{t,a}, 1 \rangle = \left\{\begin{array}{ll}
			1  & \mbox{with probability } 1-pq\\
			-1 & \mbox{with probability } pq,
		\end{array}\right.
	\end{equation}
	where we use the fact that if $\arg\max_{X_{t,a}} \langle X_{t,a}, 1 \rangle\neq 1$, it must be the case that both $X_{t, 1}$ and $X_{t,2}$ are $-1$.
	Thus, $X^\star(1)=\mathbb{E}[\arg\max_{X_{t,a}} \langle X_{t,a}, 1 \rangle]=1-2pq$, and similarly $X^\star(-1)=-1+2(1-p)(1-q)$, and hence, $\mathcal{X}=\{1-2pq, -1+2(1-p)(1-q)\}$. If $\Lambda$ decides to pick $x_t=1-2pq$, we have that $\thetat = 1$, otherwise $\thetat=-1$. This estimate $\thetat$ is then conveyed to the agent to help her pick the action.}

{\bf Algorithm Operation.} The pseudo-code is provided in Algorithm~\ref{alg:known}.\\
%\begin{itemize}
%\item
$\bullet$ First, the central  learner  calculates the function 
\begin{equation} \label{eq11}
	X^{\star}(\theta)= \mathbb{E}_{\{x_a: x_a\sim P_a\}}[\arg\max_{x\in\{x_a: a\in A\}} \langle x, \theta \rangle ],
\end{equation}
and creates the action set 
$\mathcal{X}=\{ X^{\star}(\theta)| \theta\in \Theta\}$ that algorithm $\Lambda$ is going to use. \\
%Note that this operation (calculation of $\mathcal{X}$) happens only once.\\
%\item
$\bullet$  At each time $t$, based on past history,  $\Lambda$  decides on a next action  $x_t\in \mathcal{X}$.
%the central learner 
%uses $\Lambda$ to decide  what is the new update $\hat{\theta}_t$ of the estimated parameter vector. $\Lambda$ at time $t$, decides on a next action 
%vector $x_t\in \mathcal{X}$. This action vector is not send to the agent. It is only used by
The central learner uses $x_t$  to calculate the new update
$\thetat=X^{-1}(x_t)$, where $X^{-1}$ is the inverse of $X^\star$ (see Section~\ref{sec:notation}).\\
%$\hat{\theta}_t$ to be any vector that satisfies 
% satisfies $X^*(\hat{\theta}_t)= x_t$.
%  That is, \\
%\item
$\bullet$  The  agent receives $\thetat$ from the central learner, observes her context,  plays an action $a_t=\arg\max_{a\in \mathcal{A}}\langle X_{t,a},\thetat \rangle$, and observes the reward $r_t$. 
She then quantizes the reward using a stochastic quantizer SQ$_1$  (see Section~\ref{sec:notation}), and communicates the outcome using one bit to the central learner. \\
%\item
$\bullet$ The { learner} provides the quantized reward as input to $\Lambda$. Note that $\Lambda$ is oblivious to what actions are actually played; it treats the received reward as corresponding to the action $x_t$ it had decided. 
%\end{itemize}

\begin{algorithm} 
	\caption{Communication efficient  for contextual linear bandits with known distribution}\label{alg:known}
	\begin{algorithmic}[1]
		\State Input: an algorithm ${\Lambda}$ for one context case, underlying set of actions $\mathcal{X}$, and time horizon $T$.
		\State Initialize: $X^{\star}(\theta)= \mathbb{E}_{\{x_a: x_a\sim P_a\}}[\arg\max_{x\in\{x_a: a\in A\}} \langle x, \theta \rangle ], \mathcal{X}=\{X^{\star}(\theta)|\theta\in \Theta\}$ {$,\hat{r}_0=0$}.
		\State Let $X^{-1}$ be an inverse of $X^{\star}$.
		\For{$t = 1:T$} 
		\State \textbf{Central learner}:
		\State \qquad Receive $\hat{r}_{t-1}$ and provide it to $\Lambda$.
		\State \qquad $\Lambda$, using the history 
		$(x_1,\hat{r}_1,...,x_{t-1},\hat{r}_{t-1})$, selects $x_t$.
		\State \qquad Send $\hat{\theta}_t=X^{-1}(x_t)$ to agent.
		
		\State \textbf{Agent}:
		\State \qquad Receive $\hat{\theta}_{t}$ from the central learner.
		\State \qquad Observe context realization $\{X_{t,a}\}_{a\in \mathcal{A}}$. 
		\State \qquad Pull arm $a_t=\arg\max_{a\in \mathcal{A}}\langle X_{t,a}, \hat{\theta}_{t}\rangle $ and receive reward $r_t$.
		\State \qquad Send $\hat{r}_t=\text{SQ}_1(r_t)$ to the central learner using 1-bit.
		\EndFor
	\end{algorithmic}
\end{algorithm}

The following theorem proves that Algorithm~\ref{alg:known} achieves
a regret  $R_T(\Lambda) + O(\sqrt{T\log T})$, where $R_T(\Lambda)$ is the regret of $\Lambda$ in (\ref{eq9}).  
Hence, if $\Lambda$ satisfies the best known regret bound of $O(d\sqrt{T\log T})$, e.g., LinUCB, Algorithm~\ref{alg:known} achieves a regret of $O(d\sqrt{T\log T})$.  The theorem holds under the mild set of assumptions that we stated in Section~\ref{sec:notation}.
\begin{theorem}\label{thm:known}
	Algorithm~\ref{alg:known} uses {$1$ bit per reward and $0$ bits per context}. Under Assumption~\ref{assump:1}, it achieves a regret $R_T=R_T(\Lambda)+O(\sqrt{T\log T})$ with probability at least $1-\frac{1}{T}$.
\end{theorem}
{\bf Proof outline.} The complete proof is available in App.~\ref{app:known}. We next provide a short outline. 
%The key observations in our proof are the following. \added{
	From the definition of $X^\star$ in (\ref{eq11}), we notice the following. Recall that  the distributed agent receives
	$\thetat$ from the central learner,  and pulls the best action for  this $\thetat$, i.e., $a_t=\arg\max_{a\in \mathcal{A}}\langle X_{t,a},\thetat \rangle$.
	We show that conditioned on $x_t$, the associated vector $X_{t,a_t}$  is an unbiased estimate of $x_t$ with a small variance.  Given this, we prove  that $\rt$ satisfies \eqref{eq10}, and
	{thus the rewards observed by $\Lambda$ are generated according to a linear bandit model with unknown parameter that is the same as $\thetas$.}
	
	We next decompose the difference $R_T-R_T(\Lambda)$ to two terms: $\Sigma_T=\sum_{t=1}^T \langle \arg\max_{X_{t,a}} \langle X_{t,a},\thetas\rangle,\thetas \rangle - \langle x_t,\thetas\rangle$ and $\Sigma_T'=\sum_{t=1}^T  \langle \arg\max_{X_{t,a}}\langle X_{t,a},\thetat \rangle,\thetas \rangle - \max_{x\in \mathcal{X}} \langle x,\thetas\rangle$. To bound the first term, we show that the unbiasdness property together with Assumption~\ref{assump:1} implies that $\Sigma_T$ is a martingale with bounded difference. This implies that $|\Sigma_T|=O(\sqrt{T\log T})$ with high probability. To bound $\Sigma'_T$, we first show that $\arg\max_{x\in \mathcal{X}}\langle x,\thetas \rangle=X^\star(\thetas)$ (we note that this is why the algorithm converges to $\thetat$ that is equal to, or results in the same expected reward as,  $\thetas$). Then, following a similar approach, we can show that $\Sigma_T'$ is a martingale with bounded difference which implies that $|\Sigma_T'|=O(\sqrt{T\log T})$ with high probability. \hfill{$\square$}

	%   {\bf Single Context Linear Bandits.} A direct corollary of Theorem~\ref{thm:known} is that, for the case where there is a single context, and thus, $X_{t,a}=a_t$, the agents do not need to convey to the central learner the action they decide to play - they simply need to convey one bit for the rewards. Note also that in this case the central learner does not need to know any distribution as well, as the actions do not come from a distribution. \textcolor{cyan}{CF: Does this extend to MAB as well perhaps, that we may not need to send to the central learner what is the action played?}

	{\bf Downlink Communication.} Note that in our setup %we enable unlimited downlink communication:
	we assume that the central learner does not have any communication constraints when communicating with the distributed agents. Yet our algorithm makes frugal use of this ability: the learner only sends the updated parameter vector  $\hat{\theta}_t$. 
	We can quantize $\hat{\theta}_t$ without performance loss if the downlink were also communication constrained using $\approx 5d$ bits and an approach similar to the one in Algorithm~\ref{alg:unknown}
	- yet we do not expand on this in this paper, as our focus is in minimizing  uplink communication costs.
	%, and in particular the communication of the context.
	%     \textcolor{cyan}{CF: Is this something we could easily provide in the extended materials?}\lin{we should put it in appendix.}
	
	{\bf Operation Complexity.} The main complexity that our algorithm adds beyond the complexity of $\Lambda$, is the computation of the function $X^\star$. Depending on the distributions $\mathcal{P}_a$, this can be calculated in closed form efficiently. For example, for $d=1, \theta>0$, we have that $X^\star(\theta)$ is the expectation of the maximum of multiple random variables, i.e., $X^\star(\theta)=\mathbb{E}_{x_a\sim \mathcal{P}_a}[\max_{a\in \mathcal{A}} x_a]$, which can be computed/approximated efficiently if the distributions $\mathcal{P}_a$ are given in a closed form.

	{\bf Imperfect Knowledge of Distributions.} Since we only use the distributions to calculate  $X^\star$, imperfect knowledge of distribution only affects us in the degree that  it affects the calculation of  $X^\star$. Suppose that  we have an estimate $\tilde{X}^\star$ of  $X^\star$  that satisfies 
	\begin{equation}\label{eq:mis}
		\sup_{\theta \in \Theta} \|X^\star(\theta)-\tilde{X}^\star(\theta)\|_2\leq \epsilon.
	\end{equation}
	%This estimate $\tilde{X}(\theta)$ could have been obtained from an estimate of the distributions $\mathcal{P}_a$ or by directly estimating $X^\star(\theta)$.
	Using Theorem~\ref{thm:known} we prove  in App.~\ref{app:known} the following corollary.
	\begin{corollary} \label{cor1}
		Suppose we are given $\tilde{X}^\star$ that satisfies \eqref{eq:mis}. Then, there exists an algorithm $\Lambda$ for which Algorithm~\ref{alg:known} achieves $R_T=\tilde{O}(d\sqrt{T}+\epsilon T\sqrt{d})$ with probability at least $1-\frac{1}{T}$.
	\end{corollary}
	
	%  \textcolor{cyan}{Add mispecification and any examples/cases where  we know approximately the distribution but can still achieve a reasonable performance.}

	%\textcolor{cyan}{We may remove this remark as well
		{\bf Privacy.} Our result may be useful for applications beyond communication efficiency; indeed, the context may contain private information (e.g., personal preferences, financial information, etc); use of our algorithm enables to not share this private information at all with the central learner, without impeding the learning process. Surprisingly, work in \cite{zheng2020locally}, motivated from privacy considerations, has     shown that if an agent adds a small amount of zero mean noise to the true context before sending it to the central learner, this can severely affect the regret in some cases - and yet our algorithm essentially enables to ``guess'' the context with no regret penalty if the distributions are known. Although adding a zero mean noise to the observed feature vector conveys an unbiased estimate of the observation, the difference between this and our case is technical and mainly due to the fact that the unbiasdness is required to hold conditioned on the learner observation (noisy context); see App.~\ref{app:known} for more details.

\section{Contextual Linear Bandits with Unknown Context Distribution}
\label{sec:dbits}

We now consider the case where the learner does not know the context distributions, and thus Algorithm~\ref{alg:known} that uses zero bits for the context cannot be applied. In this case, related literature conjectures a lower bound of $\Omega(d)$ \cite{wang2019distributed, zhang2013information} -- which is discouraging since it is probably impossible to establish an algorithm with communication logarithmically depending on $d$. Additionally,  in practice we use $32d$ bits to convey full precision values - thus this conjecture indicates that in practice
we may not be able to achieve order improvements in terms of bits communicated, without performance loss. 

\iffalse
In fact, it is conjectured in the literature  that with no knowledge of the distribution, a lower bound on communication is $\Omega(d)$. {This is discouraging since it is probably impossible to establish an algorithm with communication logarithmically depending on $d$. Although in theory we need infinite bits to communicate full precision values, people may use only $32d$ bits to represent real numbers with good precision, hence, requiring $32d$ for communicating $X_{t,a}$. Thus this conjecture indicates that, at least in practice, we may not be able to achieve order improvements in terms of bits communicated, without performance loss.}
They conjectured a lower bound Omega(d) -- this is discouraging since it is probably impossible to establish an algorithm with communication logarithmically depending on d. Nevertheless, the existing best algorithms take a naive approach to communicating the context vectors and costs d log T bits per context, which goes to infinity as T goes to infinity. We thus seek to improve this bound and communicate only a constant number of bits per dimension per context. On the other hand, in practice people may use only 32 bits to represent arbitrarily precision real numbers, we will establish an algorithm that uses much less number of bits than this for the context communication."
\fi

In this section, we provide Algorithm~\ref{alg:unknown} that uses $\approx 5d$ bits per context and achieves (optimal) regret $R_T=O(d\sqrt{T\log T})$. %Although this algorithm still uses $O(d)$ bits,
We believe Algorithm~\ref{alg:unknown} is interesting for two reasons:\\
1. In theory, we need an infinite number of bits  to convey full precision values- we prove that a constant number of bits per dimension per context is sufficient. Previously best-known algorithms use $O(d \log T )$ bits per context,
which goes to infinity as $T$ goes to infinity. Moreover, these algorithms  require exponential complexity \cite{lattimore2020bandit} while ours is computationally efficient. \\
2. In practice, especially for large values of $d$, reducing the number of bits conveyed from $32d$ to $\approx 5d$ is quite significant - this is a reduction by a factor of six, which implies six times less communication.

{\bf Main Idea.}
The intuition behind Algorithm~\ref{alg:unknown} is the following.
The central learner  is going to use {an estimate of the} $d\times d$ least-squares matrix ${V}_t=\sum_{i=1}^t X_{i,a_i}X_{i,a_i}^T$ to update her estimates for the parameter vector $\thetas$. Thus,  when quantizing the vector  $X_{t,a}$, we want to make sure that not only this vector is conveyed with sufficient accuracy, but also that the learner can calculate the matrix ${V}_t$ accurately. In particular, we would like the central learner to be able to calculate an unbiased estimator for each entry of $X_{t,a}$ and each entry of the matrix ${V}_t$. 
Our algorithm achieves this by  quantizing the feature vectors $X_{t,a_t}$, and also  the diagonal (only the diagonal)  entries of the least squares  matrix ${V}_t$. We prove that by doing so, with only $\approx 5d$ bits we can provide an unbiased estimate and guarantee an $O(\frac{1}{\sqrt{d}})$ quantization error for each entry in the matrix almost surely.

{\bf Quantization Scheme.} We here describe the proposed quantization scheme.\\
$\bullet$ {\em To quantize $X_{t,a_t}$:} Let $m\triangleq\ceil{\sqrt{d}}$.
We first send the sign of each coordinate of $X_{t,a_t}$ using $d$ bits, namely, we send the vector  $s_t=X_{t,a_t}/|X_{t,a_t}|$.
To quantize the magnitude $|X_{t,a_t}|$, we scale
each coordinate of $|X_{t,a_t}|$ by $m$ and quantize it using a Stochastic Quantizer (SQ)\footnote{As described in (\ref{eq3}) in Section~\ref{sec:notation}, SQ maps value $x$ to an integer value, namely	$\floor{x}$  with probability  $\ceil{x}-x$ and $\ceil{x}$  with probability  $x-\floor{x}$.}  with $m+1$ levels in the interval $[0,m]$. Let  $X_t\triangleq\text{SQ}_{m}({m |X_{t,a_t}|})$
%$\bullet$ From the resulting values, we keep only the 
denote the resulting SQ outputs, we note that $X_t$ takes  non-negative integer values and lies in a norm-1 ball of radius $2d$  (this holds since the original  vector lies in a norm-2 ball of radius $1$ and the error in each coordinate is at most $1/m$). That is, it holds that $X_t\in \mathcal{Q}=\{x\in \mathbb{N}^{d}| \|x\|_1\leq 2d \}$. We then use any enumeration  $h:\mathcal{Q}\to [|\mathcal{Q}|]$ of this set to encode $X_t$ using $\log (|\mathcal{Q}|)$ bits.\\ 
%$\bullet$ 
% For the encoding part, we use an enumeration of the set $\mathcal{Q}=\{x\in \mathbb{N}^{d}| \|x\|_1\leq 2d \}$ denoted by $h$. In particular, $h:\mathcal{Q}\to [|\mathcal{Q}|]$ is any one-to-one function that maps from $\mathcal{Q}$ to $[|\mathcal{Q}|]$ and can be thought of as an ordering of the elements in $\mathcal{Q}$. We show that our context quantizer always outputs values in the set $\mathcal{Q}$, hence, it can be encoded using $O(\log (|\mathcal{Q}|))$ bits using the ordering given by $h$.\\
$\bullet$ {\em To quantize $X_{t,a_t}X_{t,a_t}^T$}:
Let $X_{t,a_t}^2$ denote a vector that collects the diagonal entries of $X_{t,a_t}X_{t,a_t}^T$. Let $\hat{X}_t\triangleq s_tX_t/m$ be the estimate of $X_{t,a_t}$ that the central learner retrieves. Note that 
$\hat{X}_t^2$ is not an unbiased estimate of  $X_{t,a_t}^2$;
however,  $(X_{t,a_t}^2-\hat{X}_t^2)_i\leq 3/m$ for all coordinates $i$ (proved in App.~\ref{app:unknown}).
Our scheme simply conveys  the difference $X_{t,a_t}^2-\hat{X}_t^2$ with $1$ bit per coordinate using a SQ$_1^{[-3/m,3/m]}$ quantizer.
%$X_{t,a_t}^2-\hat{X}_t^2$
%although
%$X_{t,a_t}$
%can be close to $X_{t,a_t}^2$, it
%is not an unbiased estimate of $X_{t,a_t}^2$.

The central learner and distributed agent operations are presented in Algorithm~\ref{alg:unknown}.

%\textcolor{cyan}{CF: In this section, are we using norm-1 or norm-2?}\\
{\bf Example 2.} Consider the case where $d=5$. Then each coordinate of $|X_{t,a_t}|$ is scaled by $3$ and quantized using SQ$_3$ to one of the values $0,1,2,3$ to get $X_t$. The function $h$ then maps the values for $X_t$ that satisfy the $\|X_t\|_1\leq 10$ to a unique value (a code) in the set $[|\mathcal{Q}|]$. For instance the value $3.\mathbf{1}$ is not given a code, where $\mathbf{1}$ is the vector of all ones. However, note that for $|X_{t,a_t}|$ to be mapped to $3.\mathbf{1}$, we must have $3|(X_{t,a_t})_i|\geq 2$ for all coordinates $i$, which cannot happen since it implies that $\|X_{t,a_t}\|_2\geq 2\sqrt{5/6}>1$ which contradicts Assumption~\ref{assump:1}.

%\textcolor{cyan}{CF: I do not think we need to include the algorithm operation.}
\iffalse
{\bf Algorithm Operation.} 
The central learner and distributed agent operations are presented in Algorithm~\ref{alg:unknown}, and proceed as follows:\\
$\bullet$ At time $t$, the {\bf distributed agent} receives an estimate of the unknown parameter $\theta_*$, denoted by $\thetat$, from the central learner.
She also locally observes her context realizations  $\{X_{t,a}\}_{a\in \mathcal{A}}$. Then, she greedily pulls the best action for $\thetat$, namely, $a_t=\arg\max_{a\in \mathcal{A}}\langle X_{t,a},\thetat \rangle$, and receives a reward $r_t$.\\
$\bullet$ The distributed agent  quantizes $r_t$ using SQ$_1$ and one bit; quantizes $X_{t,a_t}$ as described above;  quantizes $X_{t,a_t}^2$ (the diagonal entries of $X_{t,a_t}X_{t,a_t}^T$) similarly, and transmits the quantized values to the central learner.\\
$\bullet$ The \textbf{central learner} receives the quantized estimates of $r_t, X_{t,a_t},X_{t,a_t}^2$ which we denote $\rt, \Xt, \hat{X_t^2}$ respectively, and updates $V_t, \thetat$ as in steps~\ref{step:vt}, \ref{step:theta}.
%Then, broadcasts the estimate $\thetat$ to the distributed user.
\fi

\begin{algorithm} 
	\caption{Communication efficient for contextual linear bandits with unknown distribution}\label{alg:unknown}
	\begin{algorithmic}[1]
		\State Input: underlying set of actions $\mathcal{A}$, and time horizon $T$.
		\State $\hat{\theta}_0=0, \tilde{V}_0=0, u_0=0, m=\ceil{\sqrt{d}}$.
		\State Let $h$ be an enumeration of the set $\mathcal{Q}=\{x\in \mathbb{N}^{d}| \|x\|_1\leq 2d \}$.\label{step:h}
		\For{$t = 1:T$} 
		\State \textbf{Agent}:
		\State \qquad Receive $\hat{\theta}_{t-1}$ from the central learner.
		\State \qquad Observe context realization $\{X_{t,a}\}_{a\in \mathcal{A}}$.
		\State \qquad Pull arm $a_t=\arg\max_{a\in \mathcal{A}}\langle X_{t,a}, \hat{\theta}_{t-1}\rangle $ and receive reward $r_t$.
		\State \qquad Compute the signs  $s_t=X_{t,a_t}/|X_{t,a_t}|$ of $X_{t,a_t}$.
		\State \qquad Let $X_t=\text{SQ}_{m}({m |X_{t,a_t}|})$.
		\State \qquad {$e_t^2=\text{SQ}_{1}^{[-3/m,3/m]}(X_{t,a_t}^2-\hat{X}_t^2)$, where $\hat{X}_t=s_tX_t/m$}.
		\State \qquad {Send to the central learner $h(X_t)$, $s_t$, and $e_t^2$ using $\log_2(|\mathcal{Q}|)$, $d$, and  $d$ bits, respectively.}
		\State \qquad Send $\hat{r}_t=\text{SQ}_1(r_t)$ using 1-bit.
		
		\State \textbf{Central learner}:
		\State \qquad Receive $X_t$, $s_t$, $e_t^2$, and $\hat{r}_t$ from the distributed agent.
		\State \qquad {$\hat{X}_t=s_tX_t/m$, $\hat{X}^{(D)}_t=\hat{X}_t^2+e_t^2$}.
		\State \qquad $u_t\gets u_{t-1}+\hat{r}_t\hat{X}_t$.
		\State \qquad $\tilde{V}_t\gets \tilde{V}_{t-1}+\hat{X}_t {\hat{X}_t}^T-\text{diag}(\hat{X}_t {\hat{X}_t}^T)+\text{diag}(\hat{X}^{(D)}_t)$. \label{step:vt}
		\State \qquad $\hat{\theta}_t\gets \tilde{V}_t^{-1} u_t$. \label{step:theta}
		\State \qquad Send $\hat{\theta}_t$ to the next agent.
		\EndFor
	\end{algorithmic}
\end{algorithm}
{\bf Algorithm Performance.} Theorem~\ref{thm:unknown}, stated next,  holds under 
%The following theorem provides an upper bound on the number of bits and regret for the proposed algorithm, under some assumptions stated next. In addition to
Assumption~\ref{assump:1}  in Section~\ref{sec:notation} and  some
additional regulatory assumptions on the distributions  $\mathcal{P}_a$ provided in
Assumption~\ref{assump:2}.
\begin{assumption}\label{assump:2}
	There exist constants $c,c'$ such that for any sequence $\theta_1,...,\theta_T$, where $\theta_t$ depends only on $H_t$, with probability at least $1-\frac{c'}{T}$, it holds that
	\begin{equation} \label{eq17}
		\sum_{i=1}^t X_{i,a_i}X_{i,a_i}^T\geq \frac{ct}{d}I \quad \forall t\in [T],
	\end{equation}
	where $a_t=\arg\max_{a\in \mathcal{A}}\langle X_{t,a},\theta_t \rangle$, and $I$ is the identity matrix. 
\end{assumption}
We note that several common assumptions in the literature imply
(\ref{eq17}), for example, 
bounded eigenvalues for the covariance matrix of $X_{t,a_t}$ 
\cite{ding2021efficient, li2017provably,han2020sequential}. Such assumptions hold for a wide range of distributions, including subgaussian distribitions with bounded  density \cite{ren2020dynamic}.

\begin{theorem}\label{thm:unknown}
	Algorithm~\ref{alg:unknown} satisfies that for all $t$: $X_t\in \mathcal{Q}$; and $B_t\leq 1+\log_2(2d+1)+5.03d$ bits. Under assumptions~\ref{assump:1}, \ref{assump:2}, it achieves a regret $R_T=O(d\sqrt{T\log T})$ with probability at least $1-\frac{1}{T}$.
\end{theorem}

{\bf Proof Outline.}
To bound the number of bits $B_t$, we first bound the size of $\mathcal{Q}$ by formulating a standard counting problem: we find the number of non-negative integer solutions for a linear equation. To bound the regret $R_T$, we start by proving that our quantization scheme guarantees some desirable properties, namely, unbiasedness and $O(\frac{1}{\sqrt{d}})$ quantization error for each vector coordinate. We then upper bound the regret in terms of $\|\thetat-\thetas\|_2$ and show that this difference can be decomposed as
\begin{equation}
	\|\thetat-\thetas\|_2 = \|V_t^{-1}\|_2 (\|\sum_{i=1}^t  E_i\|_2 + (1+|\eta_i|) \| \sum_{i=1}^t e_i\|_2 + \|\sum_{i=1}^t  \etai X_{i,a_i}\|_2,
\end{equation}
where $E_t$ captures the error in estimating the matrix $X_{t,a_t}X_{t,a_t}^T$, $e_t$ is the error in estimating $X_{t,a_t}$, and $\eta_t'$ is a noise that satisfies the same properties as $\eta_t$. Using Assumption~\ref{assump:2}, we prove that $V_t^{-1}$ grows as $O(\frac{d}{t})$ with high probability, and from the unbiasdness and boundedness of all error quantities we show that they grow as $O(\sqrt{t\log t})$ with high probability. This implies that $\|\thetat-\thetas\|_2=O(d\sqrt{\frac{\log t}{t}})$, and hence, $R_T=O(d\sqrt{T\log T})$. The complete proof is provided in App.~\ref{app:unknown}. \hfill{$\square$}

{\bf Algorithm Complexity.} If we do not count the quantization operations, it is easy to see that the complexity of the rest of the algorithm is dominated by the complexity of computing $V_t^{-1}$ which can be done in $O(d^{2.373})$ \cite{alman2021refined}. For the quantization, we
%The computational complexity due to quantization is as following. We
note that each coordinate of $X_t$ can be computed in $\tilde{O}(1)$ time\footnote{Multiplication by $\sqrt{d}$ can take $O(\log d)$ time.}. {Moreover, the computation of $h(x)$ for $x\in \mathcal{Q}$ can be done in constant time with high probability using hash tables, where $h$ is the enumeration function in Step~\ref{step:h}.} Hence, the added computational complexity  is almost linear in $d$. {Although a hash table for $h$ can consume $\Omega(2^{5d})$ memory, by sacrificing a constant factor in the number of bits, we can find enumeration functions that can be stored efficiently. As an example, consider the scheme in \cite{elias1975universal} that can find an one-to-one function $h:\mathcal{Q}\to \mathbb{N}^+$ which can be stored and computed efficiently, but only gives guarantees in expectation that $\mathbb{E}[\log (h(x))]=O(d)$ for all $x\in \mathcal{Q}$.}

{\bf Downlink Communication Cost.}
Although we assume no-cost downlink communication,
as was also the case for Algorithm~\ref{alg:known}, the downlink in Algorithm~\ref{alg:unknown} is only used to send the updated parameter vector  $\hat{\theta}_t$ to the agents. If desired, these estimates can be quantized using the same method as for $X_{t,a_t}$, which (following a similar proof to that of Theorem~\ref{thm:unknown}) can be shown to not affect the order of the regret while reducing the downlink communication to $\approx 5d$ bits per iteration. 

{\bf Offloading To Agents.} For applications where the agents wish to computationally help the central learner, the central  learner may simply aggregate information to keep track of $u_t$, $\tilde{V}_t$  and broadcast these values to the agents; the estimate $\thetat$ can be calculated at each  agent.  Moving the computational load to the agents does not affect the regret order or the number of bits communicated on the uplink.

\begin{remark}
	Under the regulatory assumptions in \cite{han2020sequential}, the regret bound can be improved by a factor of $\sqrt{\log(K)/d}$, where $K=|\mathcal{A}|$ is the number of actions. However, this does not improve the regret in the worst case as the worst case number of actions  is $O(C^{d}), C>1$ \cite{lattimore2020bandit}.
\end{remark}

\appendices
% \langle \rangle
\section{Proofs and Remarks for Section~\ref{sec:0bits}: Contextual Linear Bandits with Known Context Distribution} \label{app:known}

{\bf Remark.} 
We note that, to reduce the multi-context problem to a single context problem in the case of known context distribution, the straightforward approach that replaces the actual context realizations  $\{X_{t,a}\}_{a\in \mathcal{A}}$ with the fixed set $\{ \mathbb{E}_{\mathcal{P}_a}[X_{t,a}]\}_{a\in \mathcal{A}}$, and uses this latter set as $\mathcal{X}$ in $\Lambda$, does not work and can lead to linear regret in some cases. For instance consider the case where $d=1, \mathcal{A}=\{1,2\}, X_{t,a}\in \{-1,1\}\forall a\in \mathcal{A}, \thetas=1$ and $X_{t,1}$ takes the value $1$ with probability $3/4$ and $-1$ otherwise, while $X_{t,2}$ takes the values $1,-1$ with probability $1/2$. Then clearly $\langle \mathbb{E}_{\mathcal{P}_1}[X_{t,1}],\thetas \rangle > \langle \mathbb{E}_{\mathcal{P}_2}[X_{t,2}],\thetas \rangle$, however, choosing $a_t=1\forall t\in [T]$  leads to $\mathbb{E}[R_T]\geq T/8$ since it holds that $X_{t,1}=-1,X_{t,2}=1$ with probability $1/8$. 

\subsection{Proof of Theorem~\ref{thm:known}}
\addtocounter{theorem}{-2}
\begin{theorem}
	Algorithm~\ref{alg:known} uses {$1$ bit per reward and $0$ bits per context}. Under Assumption~\ref{assump:1}, it achieves a regret $R_T=R_T(\Lambda)+O(\sqrt{T\log T})$ with probability at least $1-\frac{1}{T}$.
\end{theorem}
{\bf Proof.} It is obvious that the agent only sends $1$ bit to the learner to represent $r_t$ using SQ$_1$, hence, the algorithm uses $0$ bits per context and $1$ bit per reward. We next bound the regret of our algorithm as following. The regret can be expressed as
\begin{align}\label{eq:reg_init_known}
	R_T&= \sum_{t=1}^T \max_{a\in \mathcal{A}}\langle X_{t,a},\thetas \rangle - \langle X_{t,a_t},\thetas \rangle\nonumber \\
	&=\sum_{t=1}^T \langle \arg\max_{X_{t,a}} \langle X_{t,a},\thetas\rangle,\thetas \rangle - \langle \arg\max_{X_{t,a}}\langle X_{t,a},\thetat \rangle,\thetas \rangle\nonumber \\
	&=\sum_{t=1}^T \left(\langle \arg\max_{X_{t,a}} \langle X_{t,a},\thetas\rangle,\thetas \rangle - \langle \mathbb{E}[\arg\max_{X_{t,a}} \langle X_{t,a},\thetas\rangle|\thetas],\thetas \rangle \right)\nonumber \\
	&\qquad - \left(\langle \arg\max_{X_{t,a}}\langle X_{t,a},\thetat \rangle,\thetas \rangle - \langle \mathbb{E}[\arg\max_{X_{t,a}}\langle X_{t,a},\thetat \rangle|\thetat],\thetas \rangle\right) \nonumber \\
	&\qquad +\left(\langle \mathbb{E}[\arg\max_{X_{t,a}} \langle X_{t,a},\thetas\rangle |\thetas],\thetas \rangle-\langle \mathbb{E}[\arg\max_{X_{t,a}}\langle X_{t,a},\thetat \rangle|\thetat],\thetas \rangle \right).
\end{align}
To bound $R_T$, we bound each of the three lines in the last expression. For the second term denoted by $\Sigma_t=\sum_{i=1}^t\left(\langle \arg\max_{X_{i,a}}\langle X_{i,a},\thetai \rangle,\thetas \rangle - \langle \mathbb{E}[\arg\max_{X_{i,a}}\langle X_{i,a},\thetai \rangle|\thetai],\thetas \rangle\right)$, we have that
\begin{align}
	\mathbb{E}[\Sigma_{t+1}|\Sigma_{t}]&=\Sigma_{t}+\mathbb{E}\left[\langle \arg\max_{X_{t,a}}\langle X_{t,a},\thetat \rangle,\thetas \rangle - \langle \mathbb{E}[\arg\max_{X_{t,a}}\langle X_{t,a},\thetat \rangle|\thetat],\thetas \rangle |\Sigma_t \right]\nonumber \\
	&=\Sigma_{t}+\mathbb{E}\left[\mathbb{E}\left[\langle \arg\max_{X_{t,a}}\langle X_{t,a},\thetat \rangle,\thetas \rangle - \langle \mathbb{E}[\arg\max_{X_{t,a}}\langle X_{t,a},\thetat \rangle|\thetat],\thetas \rangle |\Sigma_t,\thetat \right] |\Sigma_t \right]\nonumber \\
	&=\Sigma_{t}+\mathbb{E}\left[\mathbb{E}\left[\langle \arg\max_{X_{t,a}}\langle X_{t,a},\thetat \rangle,\thetas \rangle - \langle \mathbb{E}[\arg\max_{X_{t,a}}\langle X_{t,a},\thetat \rangle|\thetat],\thetas \rangle |\thetat \right] |\Sigma_t \right]\nonumber \\
	&=\Sigma_{t}+\mathbb{E}\left[\langle \mathbb{E}[\arg\max_{X_{t,a}}\langle X_{t,a},\thetat \rangle|\thetat],\thetas \rangle - \langle \mathbb{E}[\arg\max_{X_{t,a}}\langle X_{t,a},\thetat \rangle|\thetat],\thetas \rangle |\Sigma_t \right]\nonumber \\
	&=\Sigma_t.
\end{align}
We also have that 
\begin{align}
	|\Sigma_{t}-\Sigma_{t-1}|&\leq \|\langle \arg\max_{X_{t,a}}\langle X_{t,a},\thetat \rangle,\thetas \rangle - \langle \mathbb{E}[\arg\max_{X_{t,a}}\langle X_{t,a},\thetat \rangle],\thetas \rangle \|\|\thetas\|\nonumber \\
	&\leq \|\langle \arg\max_{X_{t,a}}\langle X_{t,a},\thetat \rangle,\thetas \rangle\|+\|\langle \mathbb{E}[\arg\max_{X_{t,a}}\langle X_{t,a},\thetat \rangle],\thetas \rangle\|\nonumber \\
	&\leq \|\arg\max_{X_{t,a}}\langle X_{t,a},\thetat \rangle\| \|\thetas \|+\| \mathbb{E}[\arg\max_{X_{t,a}}\langle X_{t,a},\thetat \rangle]\| \|\thetas \|\leq 2.
\end{align}
Hence, $\Sigma_{t}$ is a martingale with bounded difference. By Azuma–Hoeffding inequality \cite{wainwright2019high}, we have that $\|\Sigma_T\|\leq C\sqrt{T\log T}$ with probability at least $1-\frac{1}{2T}$. Similarly, the first line in \eqref{eq:reg_init_known} is a martingale with bounded difference, hence, the following holds with probability at least $1-\frac{1}{2T}$
\begin{equation}
	\left|\sum_{t=1}^T \langle \arg\max_{X_{t,a}} \langle X_{t,a},\thetas\rangle,\thetas \rangle - \langle \mathbb{E}[\arg\max_{X_{t,a}} \langle X_{t,a},\thetas\rangle|\thetas],\thetas \rangle \right|\leq C\sqrt{T\log T}.
\end{equation}
By substituting in \eqref{eq:reg_init_known} and using the union bound we get that the following holds with probability at least $1-\frac{1}{T}$
\begin{align}\label{eq:reg_sec_known}
	R_T&\leq C\sqrt{T\log T}+\sum_{t=1}^T\left(\langle \mathbb{E}[\arg\max_{X_{t,a}} \langle X_{t,a},\thetas\rangle|\thetas],\thetas \rangle-\langle \mathbb{E}[\arg\max_{X_{t,a}}\langle X_{t,a},\thetat \rangle|\thetat],\thetas \rangle \right)\nonumber \\
	&= C\sqrt{T\log T}+\sum_{t=1}^T \langle X^*(\thetas),\thetas \rangle-\langle X^*(\thetat),\thetas \rangle \nonumber \\
	&= C\sqrt{T\log T}+\sum_{t=1}^T \langle X^*(\thetas),\thetas \rangle-\langle X_t,\thetas \rangle.
\end{align}
We also have by definition of $X^*(\thetas)$ that for any given $\theta, \thetas$
\begin{align}
	\langle X^*(\thetas),\thetas \rangle &= \mathbb{E}[\max_{X_{t,a}} \langle X_{t,a},\thetas\rangle]\nonumber \\
	&\geq \mathbb{E}[\langle \arg\max_{X_{t,a}} \langle X_{t,a},\theta\rangle,\thetas\rangle] = \langle X^*(\theta),\thetas \rangle.
\end{align}
Hence, we have that $\max_{X\in \mathcal{X}}\langle X,\thetas \rangle = \langle X^*(\thetas),\thetas \rangle$. By substituting in \eqref{eq:reg_sec_known}, we get that
\begin{align}
	R_T\leq  C\sqrt{T\log T}+\sum_{t=1}^T \max_{X\in \mathcal{X}}\langle X,\thetas \rangle-\langle X_t,\thetas \rangle = C\sqrt{T\log T}+R_T(\Lambda),
\end{align}
where $R_T(\Lambda)$ is the regret of the subroutine $\Lambda$.\hfill{$\square$}

\subsection{Proof of Corollary~\ref{cor1}}
\addtocounter{corollary}{-1}
\begin{corollary}
	Suppose we are given $\tilde{X}^\star$ that satisfies \eqref{eq:mis}. Then, there exists an algorithm $\Lambda$ for which Algorithm~\ref{alg:known} achieves $R_T=\tilde{O}(d\sqrt{T}+\epsilon T\sqrt{d})$ with probability at least $1-\frac{1}{T}$.
\end{corollary}
{\bf Proof.}   $\Lambda$ assumes that the reward $r_t$ is generated according to $\langle \tilde{X}^\star(\thetat),\thetas \rangle+\eta_t$, while it is actually generated according to
\begin{equation}
	r_t=\langle X^\star(\thetat),\thetas \rangle +\eta_t= \langle \tilde{X}^\star(\thetat),\thetas \rangle +\eta_t+f(\thetat),
\end{equation}
where $f(\thetat)=\langle X^\star(\thetat)-\tilde{X}^\star(\thetat),\thetas \rangle$. We have that
\begin{equation}
	|f(\thetat)| \leq \|X^\star(\thetat)-\tilde{X}^\star(\thetat)\|\|\thetas\|\leq \epsilon.
\end{equation}
Hence, the rewards follow a misspecified linear bandit model \cite{lattimore2020bandit}. It was shown in \cite{lattimore2020bandit} that for the single context case, there is an algorithm $\Lambda$ that achieves $R_T(\Lambda)=\tilde{O}(d\sqrt{T}+\epsilon T)$ with probability at least $1-\frac{1}{T}$. The corollary follows from Theorem~\ref{thm:known} by noting that $R_T(\Lambda)$ is defined based on the true $X^\star$ as in \eqref{eq9}. \hfill{$\square$}
%========================================
\section{Proofs of Section~\ref{sec:dbits}: Contextual Linear Bandits with Unknown Context Distribution}\label{app:unknown}
\subsection{Proof of Theorem~\ref{thm:unknown}}
\begin{theorem}
	Algorithm~\ref{alg:unknown} satisfies that for all $t$: $X_t\in \mathcal{Q}$; and $B_t\leq 1+\log_2(2d+1)+5.03d$ bits. Under assumptions~\ref{assump:1}, \ref{assump:2}, it achieves a regret $R_T=O(d\sqrt{T\log T})$ with probability at least $1-\frac{1}{T}$.
\end{theorem}
{\bf Proof.} We start by proving some properties about the quantized values $\hat{X}_t,\rt, \hat{X^2_t}$. We first note that be definition of SQ, we have that
\begin{align}
	&m|(\hat{X}_t-X_{t,a_t})_j|\leq 1.
\end{align}
Hence, 
\begin{align}
	|(\hat{X}_t^2-X_{t,a_t}^2)_j|&= |(\hat{X}_t^2-(\hat{X}_t+X_{t,a_t}-\hat{X}_t)^2)_j|=|(2(X_{t,a_t}-\hat{X}_t)\hat{X}_t+(X_{t,a_t}-\hat{X}_t)^2)_i|\nonumber \\
	&\leq 2|(X_{t,a_t}-\hat{X}_t)_i| |(\hat{X}_t)_i|+|((X_{t,a_t}-\hat{X}_t)^2)_i|\leq \frac{2}{m}+\frac{1}{m^2}\leq \frac{3}{m}.
\end{align}
We also have that
\begin{align}
	\mathbb{E}[\hat{X}^{(D)}_t|X_{t,a_t}^2]&=\mathbb{E}[\hat{X}_t^2+e_t^2|X_{t,a_t}^2]=\mathbb{E}[\mathbb{E}[\hat{X}_t^2+e_t^2|X_{t,a_t}^2,\hat{X}_t]|X_{t,a_t}^2]\nonumber \\
	&=\mathbb{E}[\hat{X}_t^2+\mathbb{E}[e_t^2|X_{t,a_t}^2-\hat{X}_t^2]|X_{t,a_t}^2]=\mathbb{E}[\hat{X}_t^2+X_{t,a_t}^2-\hat{X}_t^2|X_{t,a_t}^2]=X_{t,a_t}^2.
\end{align}
In summary, from this and the definition of SQ, we get that
\begin{align}\label{eq:app_unbias}
	&m|(\hat{X}_t-X_{t,a_t})_j|\leq 1, \mathbb{E}[\hat{X}_t|X_{t,a_t}]=X_{t,a_t}\nonumber \\
	&m|\hat{X}^{(D)}_t-X_{t,a_t}^2|\leq 3, \mathbb{E}[\hat{X}^{(D)}_t|X_{t,a_t}^2]=X_{t,a_t}^2\nonumber \\
	&|\rt-r_t|\leq 1, \mathbb{E}[\rt|r_t]=r_t
\end{align}

We next show that $X_t\in \text{dom}(h)$. By definition of SQ, we have that $X_t\in \mathbb{N}^d$. We also have that
\begin{align}
	\|X_t\|_1&= \sum_{i=1}^d m|(X_{t,a_t})_i|\leq 1+\floor{\sqrt{d}|(X_{t,a_t})_i|}\leq  d+\sum_{i=1}^d \floor{\sqrt{d}|(X_{t,a})_i|}^2\nonumber \\
	&\leq d+d\|X_{t,a_t}\|^2\leq 2d.
\end{align}
Therefore, we have that $X_t\in \mathcal{Q}=\text{dom}(h)$.

We next show the upper bound on the number of bits $B_t$. We have that $\rt$ uses $1$ bit, the sign vector $s_t$ uses $d$ bits, $e_t^2$ uses $d$ bits and $X_t$ uses $\log(|\mathcal{Q}|)$ bits. We bound $|\mathcal{Q}|$ as follows. The number of non-negative solutions for the equation $\|a\|_1=x$ for $a\in \mathbb{N}^d, x\in \mathbb{N}$ is $\binom{d+x-1}{x}\leq \binom{d+x}{x}=\binom{d+x}{d}$, hence,
\begin{align}
	|\mathcal{Q}|\leq (2d+1)\binom{3d}{d}\leq (2d+1)(e\frac{3d}{d})^{d}.
\end{align}
Hence, we have that
\begin{equation}
	B_t\leq 1+\log(2d+1)+(2+\log(3e))d.
\end{equation}

We next show the regret bound. We start by bounding the regret in iteration $t$ by the distance between $\theta_\star, \hat{\theta}_{t-1}$. From step 7 of Algorithm~\ref{alg:unknown}, we have that $\langle X_{t,a_t},\hat{\theta}_{t-1} \rangle \geq \langle X_{t,a},\hat{\theta}_{t-1}\rangle \forall a\in \mathcal{A}$, hence, we have that
\begin{align}\label{eq:reg_iter}
	\max_{a\in \mathcal{A}} \langle X_{t,a},\theta_\star \rangle - \langle X_{t,a_t},\theta_\star \rangle &\leq \max_{a\in \mathcal{A}} \langle X_{t,a}-X_{t,a_t},\theta_\star \rangle-\max_{a\in \mathcal{A}} \langle X_{t,a}-X_{t,a_t},\hat{\theta}_{t-1} \rangle\nonumber \\
	&\leq \max_{a\in \mathcal{A}} \|X_{t,a}-X_{t,a_t}\| \|\theta_\star -\hat{\theta}_{t-1}\|\leq 2 \|\theta_\star -\hat{\theta}_{t-1}\|.
\end{align}
We next bound the distance $\|\theta_\star -\hat{\theta}_{t-1}\|$. Let us denote $e_t =  \hat{X}_t-X_{t,a_t}, \hat{\eta}_t=\eta_t+(\hat{r}_t-r_t), E_t=X_{t,a_t}X_{t,a_t}^T - (V_t-V_{t-1})$.
We have that
\begin{align}\label{eq:theta_init}
	\|\theta_\star-\hat{\theta}_t\| &= \|\thetas-V_t^{-1}\sum_{i=1}^t \ri \Xhi\| = \|\thetas - V_t^{-1}\sum_{i=1}^t  (X_{i,a_i}X_{i,a_i}^T \thetas+r_ie_i+\etai X_{i,a_i}+\etai e_i)\|	\nonumber \\
	& = \|V_t^{-1} \sum_{i=1}^t  (E_i \thetas +r_ie_i+\etai X_{i,a_i}+\etai e_i)\|\nonumber \\
	&\leq \|V_t^{-1}\| (\|\sum_{i=1}^t  E_i\| + (|r_i|+|\eta_i|) \| \sum_{i=1}^t e_i\| + \|\sum_{i=1}^t  \etai X_{i,a_i}\|\nonumber \\
	&\leq \|V_t^{-1}\| (\|\sum_{i=1}^t  E_i\| + (1+|\eta_i|) \| \sum_{i=1}^t e_i\| + \|\sum_{i=1}^t  \etai X_{i,a_i}\|.
\end{align}
We next bound each of the values in the last expression. 
As $\eta_i$ is subgaussian we have that with probability at least $1-\frac{1}{5T^2}$, we have that $|\eta_i|\leq C \log T \forall i\in [T]$. We also have that, using \eqref{eq:app_unbias}, $S^{e}_t=\sum_{i=1}^t e_i$ is a martingale with bounded difference, hence, by Azuma–Hoeffding inequality, we get that with probability at least $1-\frac{1}{5dT^2}$ we have that $|(S^{e}_t)_j|\leq \frac{C}{\sqrt{d}} \sqrt{t\log (dT)}$; note that $|(e_t)_i|\leq \frac{1}{\sqrt{d}}$. Hence, by the union bound we get that with probability at least $1-\frac{1}{5T^2}$ we have that $\| \sum_{i=1}^t e_i\|\leq C\sqrt{t\log(dT)}$. Similarly, conditioned on $X_{1,a_1},...,X_{t,a_t}$, $\sum_{i=1}^t  \etai X_{i,a_i}$ is a martingale with bounded difference, hence, with probability at least $1-\frac{1}{5dT^2}$ we have that $|(\sum_{i=1}^t  \etai X_{i,a_i})_j|\leq C \sqrt{\sum_{i=1}^{t}(X_{i,a_i})_j^2 \log (dT)}$. Hence, with probability at least $1-\frac{1}{5T^2}$ we have that $\|\sum_{i=1}^t  \etai X_{i,a_i}\|\leq C\sqrt{\sum_{i=1}^t \|X_{i,a_i}\|^2\log(dT)}\leq C\sqrt{t\log(dT)}$. Summing up, we get that with probability at least $1-\frac{3}{5T^2}$
\begin{align}
	\|\theta_\star-\hat{\theta}_t\|\leq \|V_t^{-1}\| (\|\sum_{i=1}^t  E_i\|+C\sqrt{t\log(dT)}).
\end{align}
It remains to bound $\|V_t^{-1}\|, \|\sum_{i=1}^t  E_i\|$ which we do in the following by starting with $\|\sum_{i=1}^t  E_i\|$. We have that
\begin{align}
	E_i &= X_{i,a_i}X_{i,a_i}^T-\Xhi \Xhi^T +\text{diag}(\Xhi \Xhi^T) - \text{diag}(\hat{X}^{(D)}_t) \nonumber \\
	&= \text{diag}(\Xhi \Xhi^T) - 2 X_{i,a_i} e_i^T - e_i e_i^T - \text{diag}(\hat{X}^{(D)}_t)\nonumber \\
	& = 2 \text{diag}(X_{i,a_i} e_i^T) - 2 X_{i,a_i} e_i^T - (e_i e_i^T-\text{diag}(e_i e_i^T)) - \text{diag}(\hat{X}^{(D)}_t-X_{i,a_i}^2).
\end{align}
Hence, we have that
\begin{align}\label{eq:boundE1}
	\|\sum_{i=1}^t  E_i\| \leq &2 \|\sum_{i=1}^t \text{diag}(X_{i,a_i} e_i^T)\| + 2 \|\sum_{i=1}^t X_{i,a_i} e_i^T\| \nonumber \\
	&+ \|\sum_{i=1}^t e_i e_i^T-\text{diag}(e_i e_i^T)\| + \|\sum_{i=1}^t \text{diag}(\hat{X}^{(D)}_t-X_{i,a_i}^2)\|.
\end{align}
We have that, using \eqref{eq:app_unbias}, conditioned on $X_{1,a_1},...,X_{t,a_t}$, $\sum_{i=1}^t \text{diag}(X_{i,a_i} e_i^T)$ is a martingale with bounded difference, hence, similar to what we did before using Azuma–Hoeffding inequality and the union bound we get that with probability at least $1-\frac{1}{20T^2}$, we have that  $\|\sum_{i=1}^t \text{diag}(X_{i,a_i} e_i^T)\|\leq \frac{C}{\sqrt{d}}\sqrt{t\log(dT)}$. Similarly, with probability at least $1-\frac{1}{20T^2}$, we have that$\|\sum_{i=1}^t \text{diag}(\hat{X}^{(D)}_t-X_{i,a_i}^2)\|\leq \frac{C}{\sqrt{d}}\sqrt{t\log(dT)}$. We next turn to bounding $\|\sum_{i=1}^t X_{i,a_i} e_i^T\|$. Conditioned on $X_{1,a_1},...,X_{t,a_t}$, we have that by Azuma–Hoeffding, with probability at least $1-\frac{1}{d^2T^2}$, we have
\begin{equation}
	|(\sum_{i=1}^t X_{i,a_i} e_i^T)_{jk}|\leq \frac{C}{\sqrt{d}}\sqrt{\sum_{i=1}^t(X_{i,a_i})_j^2\log(dT)}.
\end{equation}
We notice that taking the absolute value of all elements of a matrix does not decrease its maximum eigenvalue, hence, by the union bound we have that with probability at least $1-\frac{1}{20T^2}$ we have that
\begin{align}
	\|\sum_{i=1}^t X_{i,a_i} e_i^T\| &\leq \frac{C\sqrt{\log(dT)}}{\sqrt{d}}\|\mathbf{1}[\sqrt{\sum_{i=1}^t(X_{i,a_i})_1^2},...,\sqrt{\sum_{i=1}^t(X_{i,a_i})_d^2}]\|\nonumber \\
	&\leq \frac{C\sqrt{\log(dT)}}{\sqrt{d}}\sqrt{d\sum_{i=1}^t\|X_{i,a_i}\|^2} \leq C\sqrt{t\log(dT)}.
\end{align}
To bound $\|\sum_{i=1}^t e_i e_i^T-\text{diag}(e_i e_i^T)\|$, we notice that for all elements except the diagonal we have that $\mathbb{E}[(e_i)_j(e_i)_k]=\mathbb{E}[(e_i)_j(e_i)_k|X_{t,a_t}]=\mathbb{E}[(e_i)_k|X_{t,a_t}]\mathbb{E}[(e_i)_j|X_{t,a_t}]=0, j\neq k$. Hence, it can be shown that $\sum_{i=1}^t (e_i)_j (e_i)_k$ is a martingale with bounded difference for $j\neq k$, hence, with probability at least $1-\frac{1}{20d^2T^2}$, we have that $|\sum_{i=1}^t (e_i)_j (e_i)_k|\leq \frac{C}{d}\sqrt{t\log(dT)}$. Hence, by the union bound we get that with probability at least $1-\frac{1}{20T^2}$
\begin{equation}
	\|\sum_{i=1}^t e_i e_i^T-\text{diag}(e_i e_i^T)\|\leq \frac{C\sqrt{t\log(dT)}}{d}\|\mathbf{1}\mathbf{1}^T\|\leq C\log(dT).
\end{equation}
Hence, from \eqref{eq:boundE1} and the union bound we have that with probability at least $1-\frac{1}{5T^2}$
\begin{equation}\label{eq:boundE_f}
	\|\sum_{i=1}^t  E_i\| \leq C\sqrt{t\log(dT)}.
\end{equation}
We next turn to bounding $\|V_t^{-1}\|$. We have from \eqref{eq:boundE_f}, and Assumption~\ref{assump:2}, and the union bound, the following holds with probability at least $1-\frac{2}{5T^2}$
\begin{align}
	\|V_t\|=\|\sum_{i=1}^tX_{i,a_i}X_{i,a_i}^T-E_i\|\geq \|\sum_{i=1}^tX_{i,a_i}X_{i,a_i}^T\|-\|E_i\|\geq C(\frac{t}{d}-\sqrt{t\log(dT)}).
\end{align}
Hence for $t\geq 4 \log(dT)$, we have that with probability at least $1-\frac{2}{5T^2}$, it holds that $\|V_t\|\geq C\frac{t}{2d}$, and hence,
\begin{equation}
	\|V_t^{-1}\|\leq C\frac{d}{t}.
\end{equation}
Hence, from \eqref{eq:theta_init} and the union bound, the following holds with probability at least $1-\frac{1}{T^2}$
\begin{equation}
	\|\theta_\star-\hat{\theta}_t\|\leq Cd\frac{\sqrt{\log(dT)}}{\sqrt{t}}.
\end{equation}
Therefore, from \eqref{eq:reg_iter} and the union bound again we have that the following holds with probability at least $1-\frac{1}{T}$
\begin{align}
	R_T&\leq \sum_{t=1}^T Cd\frac{\sqrt{\log(dT)}}{\sqrt{t}}\leq Cd\sqrt{\log(dT)}(1+\int_{t=1}^T\frac{1}{\sqrt{t}}dt)\nonumber \\
	&\leq 2Cd\sqrt{T\log(dT)}.
\end{align}\hfill{$\square$}

}

\newpage
\bibliographystyle{IEEEtran}
\bibliography{Refs}

\begin{thebibliography}{10}
\providecommand{\url}[1]{#1}
\csname url@samestyle\endcsname
\providecommand{\newblock}{\relax}
\providecommand{\bibinfo}[2]{#2}
\providecommand{\BIBentrySTDinterwordspacing}{\spaceskip=0pt\relax}
\providecommand{\BIBentryALTinterwordstretchfactor}{4}
\providecommand{\BIBentryALTinterwordspacing}{\spaceskip=\fontdimen2\font plus
\BIBentryALTinterwordstretchfactor\fontdimen3\font minus
  \fontdimen4\font\relax}
\providecommand{\BIBforeignlanguage}[2]{{%
\expandafter\ifx\csname l@#1\endcsname\relax
\typeout{** WARNING: IEEEtran.bst: No hyphenation pattern has been}%
\typeout{** loaded for the language `#1'. Using the pattern for}%
\typeout{** the default language instead.}%
\else
\language=\csname l@#1\endcsname
\fi
#2}}
\providecommand{\BIBdecl}{\relax}
\BIBdecl

\bibitem{auer2002using}
P.~Auer, ``Using confidence bounds for exploitation-exploration trade-offs,''
  \emph{Journal of Machine Learning Research}, vol.~3, no. Nov, pp. 397--422,
  2002.

\bibitem{mary2015bandits}
J.~Mary, R.~Gaudel, and P.~Preux, ``Bandits and recommender systems,'' in
  \emph{International Workshop on Machine Learning, Optimization and Big
  Data}.\hskip 1em plus 0.5em minus 0.4em\relax Springer, 2015, pp. 325--336.

\bibitem{sajeev2021contextual}
S.~Sajeev, J.~Huang, N.~Karampatziakis, M.~Hall, S.~Kochman, and W.~Chen,
  ``Contextual bandit applications in a customer support bot,'' in
  \emph{Proceedings of the 27th ACM SIGKDD Conference on Knowledge Discovery \&
  Data Mining}, 2021, pp. 3522--3530.

\bibitem{durand2018contextual}
A.~Durand, C.~Achilleos, D.~Iacovides, K.~Strati, G.~D. Mitsis, and J.~Pineau,
  ``Contextual bandits for adapting treatment in a mouse model of de novo
  carcinogenesis,'' in \emph{Machine learning for healthcare conference}.\hskip
  1em plus 0.5em minus 0.4em\relax PMLR, 2018, pp. 67--82.

\bibitem{awerbuch2004adaptive}
B.~Awerbuch and R.~D. Kleinberg, ``Adaptive routing with end-to-end feedback:
  Distributed learning and geometric approaches,'' in \emph{Proceedings of the
  thirty-sixth annual ACM symposium on Theory of computing}, 2004, pp. 45--53.

\bibitem{li2020multi}
F.~Li, D.~Yu, H.~Yang, J.~Yu, H.~Karl, and X.~Cheng, ``Multi-armed-bandit-based
  spectrum scheduling algorithms in wireless networks: A survey,'' \emph{IEEE
  Wireless Communications}, vol.~27, no.~1, pp. 24--30, 2020.

\bibitem{le2014sequential}
T.~Le, C.~Szepesvari, and R.~Zheng, ``Sequential learning for multi-channel
  wireless network monitoring with channel switching costs,'' \emph{IEEE
  Transactions on Signal Processing}, vol.~62, no.~22, pp. 5919--5929, 2014.

\bibitem{bouneffouf2017bandit}
D.~Bouneffouf, I.~Rish, and G.~A. Cecchi, ``Bandit models of human behavior:
  Reward processing in mental disorders,'' in \emph{International Conference on
  Artificial General Intelligence}.\hskip 1em plus 0.5em minus 0.4em\relax
  Springer, 2017, pp. 237--248.

\bibitem{matikainen2013multi}
P.~Matikainen, P.~M. Furlong, R.~Sukthankar, and M.~Hebert, ``Multi-armed
  recommendation bandits for selecting state machine policies for robotic
  systems,'' in \emph{2013 IEEE International Conference on Robotics and
  Automation}.\hskip 1em plus 0.5em minus 0.4em\relax IEEE, 2013, pp.
  4545--4551.

\bibitem{rafferty2018bandit}
A.~N. Rafferty, H.~Ying, and J.~J. Williams, ``Bandit assignment for
  educational experiments: Benefits to students versus statistical power,'' in
  \emph{International Conference on Artificial Intelligence in
  Education}.\hskip 1em plus 0.5em minus 0.4em\relax Springer, 2018, pp.
  286--290.

\bibitem{anisi2015survey}
M.~H. Anisi, G.~Abdul-Salaam, and A.~H. Abdullah, ``A survey of wireless sensor
  network approaches and their energy consumption for monitoring farm fields in
  precision agriculture,'' \emph{Precision Agriculture}, vol.~16, no.~2, pp.
  216--238, 2015.

\bibitem{novlan2013analytical}
T.~D. Novlan, H.~S. Dhillon, and J.~G. Andrews, ``Analytical modeling of uplink
  cellular networks,'' \emph{IEEE Transactions on Wireless Communications},
  vol.~12, no.~6, pp. 2669--2679, 2013.

\bibitem{hanna2022solving}
O.~A. Hanna, L.~Yang, and C.~Fragouli, ``Solving multi-arm bandit using a few
  bits of communication,'' in \emph{International Conference on Artificial
  Intelligence and Statistics}.\hskip 1em plus 0.5em minus 0.4em\relax PMLR,
  2022, pp. 11\,215--11\,236.

\bibitem{lattimore2020bandit}
T.~Lattimore and C.~Szepesv{\'a}ri, \emph{Bandit algorithms}.\hskip 1em plus
  0.5em minus 0.4em\relax Cambridge University Press, 2020.

\bibitem{shi2021federated}
C.~Shi and C.~Shen, ``Federated multi-armed bandits,'' in \emph{Proceedings of
  the 35th AAAI Conference on Artificial Intelligence (AAAI)}, 2021.

\bibitem{shahrampour2017multi}
S.~Shahrampour, A.~Rakhlin, and A.~Jadbabaie, ``Multi-armed bandits in
  multi-agent networks,'' in \emph{2017 IEEE International Conference on
  Acoustics, Speech and Signal Processing (ICASSP)}.\hskip 1em plus 0.5em minus
  0.4em\relax IEEE, 2017, pp. 2786--2790.

\bibitem{anantharam1987asymptotically}
V.~Anantharam, P.~Varaiya, and J.~Walrand, ``Asymptotically efficient
  allocation rules for the multiarmed bandit problem with multiple plays-part
  i: Iid rewards,'' \emph{IEEE Transactions on Automatic Control}, vol.~32,
  no.~11, pp. 968--976, 1987.

\bibitem{anandkumar2011distributed}
A.~Anandkumar, N.~Michael, A.~K. Tang, and A.~Swami, ``Distributed algorithms
  for learning and cognitive medium access with logarithmic regret,''
  \emph{IEEE Journal on Selected Areas in Communications}, vol.~29, no.~4, pp.
  731--745, 2011.

\bibitem{landgren2019distributed}
P.~C. Landgren, ``Distributed multi-agent multi-armed bandits,'' Ph.D.
  dissertation, Princeton University, 2019.

\bibitem{jech2008axiom}
T.~J. Jech, \emph{The axiom of choice}.\hskip 1em plus 0.5em minus 0.4em\relax
  Courier Corporation, 2008.

\bibitem{durrett2019probability}
R.~Durrett, \emph{Probability: theory and examples}.\hskip 1em plus 0.5em minus
  0.4em\relax Cambridge university press, 2019, vol.~49.

\bibitem{abbasi2011improved}
Y.~Abbasi-Yadkori, D.~P{\'a}l, and C.~Szepesv{\'a}ri, ``Improved algorithms for
  linear stochastic bandits,'' \emph{Advances in neural information processing
  systems}, vol.~24, 2011.

\bibitem{rusmevichientong2010linearly}
P.~Rusmevichientong and J.~N. Tsitsiklis, ``Linearly parameterized bandits,''
  \emph{Mathematics of Operations Research}, vol.~35, no.~2, pp. 395--411,
  2010.

\bibitem{agrawal2013thompson}
S.~Agrawal and N.~Goyal, ``Thompson sampling for contextual bandits with linear
  payoffs,'' in \emph{International conference on machine learning}.\hskip 1em
  plus 0.5em minus 0.4em\relax PMLR, 2013, pp. 127--135.

\bibitem{gray1993dithered}
R.~M. Gray and T.~G. Stockham, ``Dithered quantizers,'' \emph{IEEE Transactions
  on Information Theory}, vol.~39, no.~3, pp. 805--812, 1993.

\bibitem{zheng2020locally}
K.~Zheng, T.~Cai, W.~Huang, Z.~Li, and L.~Wang, ``Locally differentially
  private (contextual) bandits learning,'' \emph{Advances in Neural Information
  Processing Systems}, vol.~33, pp. 12\,300--12\,310, 2020.

\bibitem{wang2019distributed}
Y.~Wang, J.~Hu, X.~Chen, and L.~Wang, ``Distributed bandit learning:
  Near-optimal regret with efficient communication,'' in \emph{International
  Conference on Learning Representations}, 2019.

\bibitem{zhang2013information}
Y.~Zhang, J.~Duchi, M.~I. Jordan, and M.~J. Wainwright, ``Information-theoretic
  lower bounds for distributed statistical estimation with communication
  constraints,'' \emph{Advances in Neural Information Processing Systems},
  vol.~26, 2013.

\bibitem{ding2021efficient}
Q.~Ding, C.-J. Hsieh, and J.~Sharpnack, ``An efficient algorithm for
  generalized linear bandit: Online stochastic gradient descent and thompson
  sampling,'' in \emph{International Conference on Artificial Intelligence and
  Statistics}.\hskip 1em plus 0.5em minus 0.4em\relax PMLR, 2021, pp.
  1585--1593.

\bibitem{li2017provably}
L.~Li, Y.~Lu, and D.~Zhou, ``Provably optimal algorithms for generalized linear
  contextual bandits,'' in \emph{International Conference on Machine
  Learning}.\hskip 1em plus 0.5em minus 0.4em\relax PMLR, 2017, pp. 2071--2080.

\bibitem{han2020sequential}
Y.~Han, Z.~Zhou, Z.~Zhou, J.~Blanchet, P.~W. Glynn, and Y.~Ye, ``Sequential
  batch learning in finite-action linear contextual bandits,'' \emph{arXiv
  preprint arXiv:2004.06321}, 2020.

\bibitem{ren2020dynamic}
Z.~Ren and Z.~Zhou, ``Dynamic batch learning in high-dimensional sparse linear
  contextual bandits,'' \emph{arXiv preprint arXiv:2008.11918}, 2020.

\bibitem{alman2021refined}
J.~Alman and V.~V. Williams, ``A refined laser method and faster matrix
  multiplication,'' in \emph{Proceedings of the 2021 ACM-SIAM Symposium on
  Discrete Algorithms (SODA)}.\hskip 1em plus 0.5em minus 0.4em\relax SIAM,
  2021, pp. 522--539.

\bibitem{elias1975universal}
P.~Elias, ``Universal codeword sets and representations of the integers,''
  \emph{IEEE transactions on information theory}, vol.~21, no.~2, pp. 194--203,
  1975.

\bibitem{seide20141}
F.~Seide, H.~Fu, J.~Droppo, G.~Li, and D.~Yu, ``1-bit stochastic gradient
  descent and its application to data-parallel distributed training of speech
  dnns,'' in \emph{Fifteenth Annual Conference of the International Speech
  Communication Association}.\hskip 1em plus 0.5em minus 0.4em\relax Citeseer,
  2014.

\bibitem{alistarh2017qsgd}
D.~Alistarh, D.~Grubic, J.~Li, R.~Tomioka, and M.~Vojnovic, ``Qsgd:
  Communication-efficient sgd via gradient quantization and encoding,''
  \emph{Advances in Neural Information Processing Systems}, vol.~30, 2017.

\bibitem{mayekar2020ratq}
P.~Mayekar and H.~Tyagi, ``Ratq: A universal fixed-length quantizer for
  stochastic optimization,'' in \emph{International Conference on Artificial
  Intelligence and Statistics}.\hskip 1em plus 0.5em minus 0.4em\relax PMLR,
  2020, pp. 1399--1409.

\bibitem{hanna2021quantization}
O.~A. Hanna, Y.~H. Ezzeldin, C.~Fragouli, and S.~Diggavi, ``Quantization of
  distributed data for learning,'' \emph{IEEE Journal on Selected Areas in
  Information Theory}, vol.~2, no.~3, pp. 987--1001, 2021.

\bibitem{hanna2020distributed}
O.~A. Hanna, Y.~H. Ezzeldin, T.~Sadjadpour, C.~Fragouli, and S.~Diggavi, ``On
  distributed quantization for classification,'' \emph{IEEE Journal on Selected
  Areas in Information Theory}, vol.~1, no.~1, pp. 237--249, 2020.

\bibitem{gersho2012vector}
A.~Gersho and R.~M. Gray, \emph{Vector quantization and signal
  compression}.\hskip 1em plus 0.5em minus 0.4em\relax Springer Science \&
  Business Media, 2012, vol. 159.

\bibitem{wainwright2019high}
M.~J. Wainwright, \emph{High-dimensional statistics: A non-asymptotic
  viewpoint}.\hskip 1em plus 0.5em minus 0.4em\relax Cambridge University
  Press, 2019, vol.~48.

\end{thebibliography}

\end{document}